\documentclass{article}
\usepackage{iclr2027_conference,times}

\usepackage{amsmath,amsfonts,bm}

\def\eqref#1{equation~\ref{#1}}

\def\1{\bm{1}}

\DeclareMathAlphabet{\mathsfit}{\encodingdefault}{\sfdefault}{m}{sl}
\SetMathAlphabet{\mathsfit}{bold}{\encodingdefault}{\sfdefault}{bx}{n}

\usepackage{graphicx}
\usepackage{xcolor}
\usepackage{booktabs}
\usepackage{multirow}
\usepackage{tabularx}
\usepackage{adjustbox}
\usepackage{array}
\usepackage{enumitem}
\usepackage{capt-of}
\usepackage{hyperref}
\usepackage{url}
\usepackage[nameinlink,capitalise]{cleveref}

\definecolor{red}{rgb}{1.0,0.0,0.0}

\def\ttt{TaoTex}

\let\cite\citep
\crefname{section}{Sec.}{Secs.}
\Crefname{section}{Sec.}{Secs.}
\crefname{figure}{Fig.}{Figs.}
\Crefname{figure}{Fig.}{Figs.}
\crefname{table}{Tab.}{Tabs.}
\Crefname{table}{Tab.}{Tabs.}

\title{\ttt: Boosting Texture Detail Fidelity for Native 3D Material Generation}

\author{
Xiuchao Wu$^*$ \quad Shuichang Lai$^*$ \quad Jiangjing Lyu$^{\dagger}$
\quad Chengfei Lyu$^{\dagger}$\\Alibaba Group
}

\iclrfinalcopy 

\begin{document}

\maketitle

\begin{center}
    \centering
    \includegraphics[width=\linewidth,trim=0 0 0 0,clip]{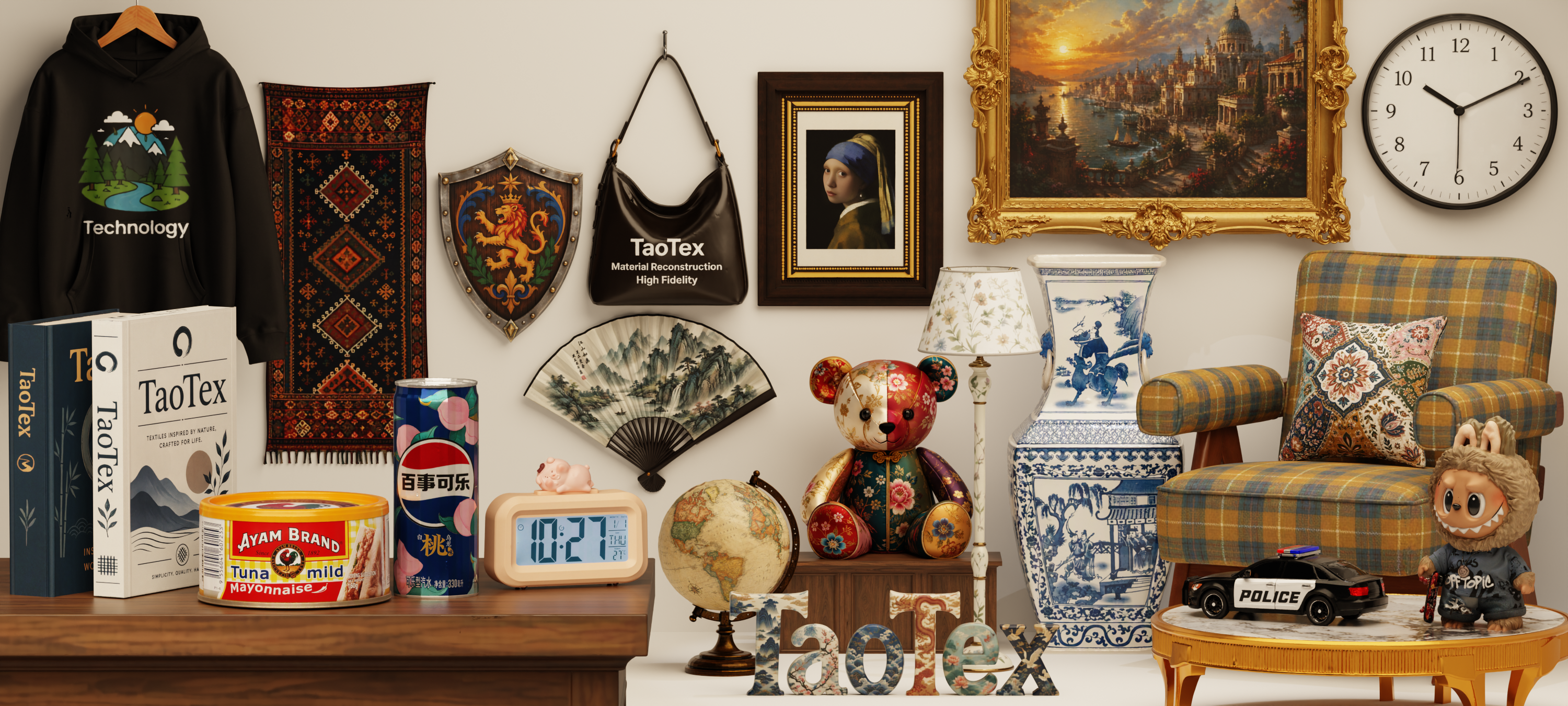}
    \captionof{figure}{\textbf{Native 3D materials generated by \ttt.} \ttt\ is capable of generating highly faithful materials across a wide range of categories, including challenging details such as text and logos. The geometries shown in the figure above are generated by TRELLIS.2~\cite{xiang2025trellis2}.}
    \label{fig:teaser}
\end{center}

\renewcommand{\thefootnote}{}
\footnotetext{$^*$Equal contribution \quad $\dagger$Corresponding author.}
\renewcommand{\thefootnote}{\arabic{footnote}}

\begin{abstract}

Recent 3D generation models can produce accurate geometries while still struggling to reconstruct detailed textures. We propose a diffusion-based native 3D material generation model \ttt, which faithfully recovers intricate textures through tailored strategies and improvements.
First, we develop a data construction agent to create high-frequency textured 3D assets to bridge the data gap in public datasets. Training with these data significantly enhances the ability of \ttt\ to recover challenging details such as text and patterns. Second, we design a multi-level feature fusion (MLFF) module to adaptively integrate local and global features of the conditional input, providing more complete texture cues for the diffusion model and thereby enhancing reconstruction fidelity. To alleviate VAE reconstruction errors, we adopt a latent-to-pixel space loss transition, further improving the pixel-level details and generation quality. Finally, we scale \ttt\ to multi-view inputs by incorporating learnable viewpoint embeddings, achieving accurate and consistent material reconstruction across views. Extensive experiments demonstrate that our method significantly outperforms existing approaches in preserving texture details in both single- and multi-view settings.

\end{abstract}
\section{Introduction}
\label{sec:intro}

Diffusion-based Image-to-3D asset generation~\cite{chang2025reconviagen, yang2024hunyuan3d, hunyuan3d2025hunyuan3d, hunyuan3d22025tencent, xiang2024structured, xiang2025trellis2} has advanced rapidly, yet existing methods struggle to faithfully reconstruct complex textures on 3D surfaces from input images, restricting their applicability to objects with simple textures rather than texture-rich categories. Current methods for generating materials on 3D assets typically operate either in view-space~\cite{hunyuan3d22025tencent, hunyuan3d2025hunyuan3d, he2025materialmvp, wu2026matmart, huang2025material, zhang2026pixtex, liang2026unitex, han2026ink3d} or directly in 3D-space~\cite{lai2026natex, xiang2024structured, xiang2025trellis2, chang2025reconviagen}. While view-space methods can leverage the priors of image generation models to assist texture synthesis, they still suffer from significant limitations. For example, texture baking may produce ghosting in overlapping regions, self-occluded areas require additional UV-space inpainting, and texture-geometry misalignment can cause projection artifacts. In contrast, generating material properties in 3D space inherently ensures global consistency and avoids the issue of self-occlusion.

\begin{figure}
\centering
\includegraphics[width=\linewidth]{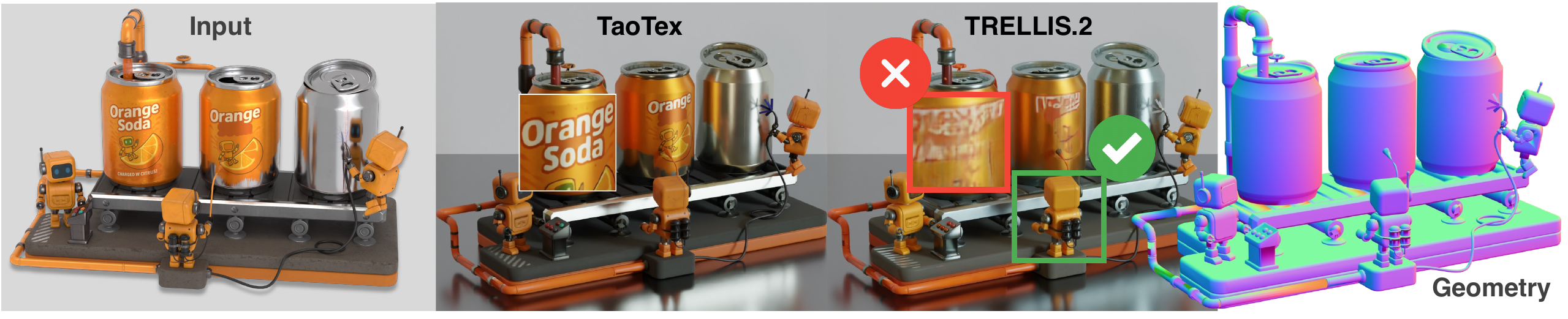}
\caption{\textbf{Texture generation and geometric cues.} In the green box, geometric variations align well with texture changes, effectively guiding appearance generation. Conversely, the red box contains feature-sparse surfaces where geometry provides little guidance for synthesizing patterns and text.
}
\label{fig:problem} 
\vspace{-0.5cm}
\end{figure}
However, existing 3D-space methods often fail to generate accurate patterns on planar or smooth surfaces. These regions lack distinctive geometric cues for the model to match with the conditional images, leading to incorrect or fragmented textures (\cref{fig:problem}). We attribute this limitation to three factors: 1) Current public 3D datasets provide insufficient high-frequency textures, leaving the mapping from intricate patterns to simple surfaces unexplored by the model. 2) Conditional image features are semantically biased and lack detailed information. 3) Latent-space training is bounded by the reconstruction accuracy of the 3D VAE. Moreover, these methods either do not support multi-view input or perform poorly on it, despite its necessity in practical applications.

To address these issues, we propose \ttt, a native material generation model with the capability to recover texture details from both single-view and multi-view inputs. For training this model, we develop a 3D asset data construction agent based on the Qwen LLM~\cite{QWEN_AI}, which enables the automatic creation of diverse 3D assets with detailed text, logo patterns, and various material properties. To better capture fine details from the conditional images, we propose a Multi-level Feature Fusion (MLFF) module that adaptively integrates both semantic and local information, allowing a more faithful reconstruction. To mitigate the accuracy degradation caused by the 3D VAE, we use a two-stage training strategy, starting with latent-space loss and then switching to pixel-space loss. Furthermore, we scale \ttt\ to multi-view inputs by introducing learnable viewpoint embeddings. Combined with our constructed dataset and trained with a low-SNR schedule, this enables accurate and consistent texture reconstruction across different views.

The primary contributions of this work are as follows:

\begin{enumerate}[label=\textbullet,leftmargin=14pt]

\item Design a novel data construction agent to create high-frequency textured 3D assets for training, boosting the ability of the native 3D material generation model to handle patterns and text.

\item Propose the MLFF module, coupled with the latent-to-pixel space loss transition strategy, enabling \ttt\ to reconstruct higher-fidelity texture details from the input images.

\item Develop a training strategy for multi-view reconstruction, incorporating learnable viewpoint embeddings, symmetric geometry data, and a low-SNR schedule, which enables \ttt\ to support both single- and multi-view inputs while achieving globally consistent and seamless texture reconstruction via a generation framework that outperforms existing leading methods.

\end{enumerate}

\section{Related Work}
\label{sec:related_work}

\subsection{View-Space Texture Generation}
Leveraging 2D diffusion priors~\cite{rombach2022high}, early methods optimize surface appearance via Score Distillation Sampling (SDS)~\cite{poole2022dreamfusion,chen2023fantasia3d,lin2023magic3d,liang2024luciddreamer,zhang2024dreammat}, which is computationally expensive and prone to over-saturation and Janus artifacts. To improve efficiency, TEXTure~\cite{richardson2023texture} and Text2Tex~\cite{chen2023text2tex} sequentially synthesize and project individual views onto meshes, but their view-by-view pipeline lacks direct cross-view interaction, accumulating inconsistencies across viewpoints.

Recent methods adopt multi-view diffusion (MVD) to improve cross-view consistency~\cite{liu2023zero,shi2023zero123pp,long2024wonder3d,li2024era3d}. Subsequent works improve texture quality through synchronized denoising, geometric conditioning, cross-view attention, and specialized projection strategies~\cite{liu2024text,zeng2024paint3d,cheng2024mvpaint,bensadoun2024meta,hunyuan3d22025tencent,hunyuan3d2025hunyuan3d,liu2025calitex}. In parallel, Ink3D~\cite{han2026ink3d} employs video diffusion for continuous orbit-scan coverage. Other recent methods extend view-space synthesis from RGB appearances to PBR material properties, including albedo, roughness, and metallic~\cite{he2025materialmvp,huang2025material,wu2026matmart}.
Despite substantial progress, view-space methods still suffer from cross-view inconsistencies in high-frequency regions, incomplete coverage under self-occlusion, and projection artifacts along depth boundaries.

\subsection{Native 3D Texture Generation}
Native 3D methods model appearance directly in geometry-aligned spaces. Early works synthesize per-face colors using adversarial objectives~\cite{siddiqui2022texturify} or continuous implicit fields~\cite{oechsle2019texture}. Modern diffusion-based approaches investigate varied representations, including colored point clouds, octree grids, UV parameterizations, and 3D Gaussian attributes~\cite{yu2023texture,liu2024texoct,yu2024texgen,liu2025texgarment,xiong2024texgaussian}, each presenting trade-offs between resolution, efficiency, and graphics compatibility. Concurrently, unified frameworks generate geometry and texture jointly~\cite{hong2023lrm,tang2024lgm}, though their structural coupling restricts independent texture optimization.

More recently, structured latent spaces have established a powerful foundation for native 3D appearance modeling. Pioneered by Trellis~\cite{xiang2024structured,xiang2025trellis2}, geometry and texture can be jointly parameterized within a unified Structured LATent (SLAT) manifold. Targeting texture synthesis on fixed shapes, NaTex~\cite{lai2026natex} introduces an explicit decoupling of geometric structure and surface color using a dual-stream autoencoder enriched with cross-latent feature aggregation. Building upon spatial sparsity and coordinate continuity, recent approaches perform diffusion on sparse voxel features or topology-agnostic texture formulations~\cite{zeng2026textrix,liang2026unitex}, expanding the generative scope from raw radiance to complete physically based material channels~\cite{yeo2026matlat}. While 3D-space methods effectively ensure global consistency and seamless surface coverage, existing frameworks fall short of resolving sharp, high-frequency appearance details.

\section{Method}
\label{sec:method}

\ttt\ is a native 3D generative model based on the o-voxel representation~\cite{xiang2025trellis2}, producing PBR materials (albedo, roughness, and metallic) from an input geometry and one or multiple images. Below, we present our data construction agent (\cref{sec:agent}), key improvements for boosting texture details (\cref{sec:scheme}), and the extension of this backbone to multi-view setting (\cref{sec:multi}).

\begin{figure*}[t]
\centering
\includegraphics[width=\linewidth]{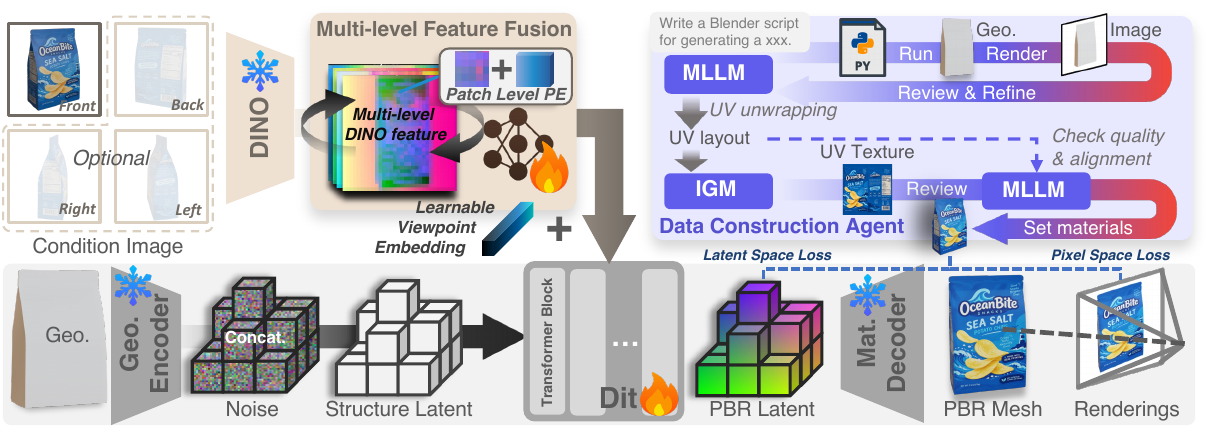}
\caption{\textbf{Method overview.} We advance native 3D material detail reconstruction across three fronts: a data agent for high-frequency asset curation, an MLFF module fusing multi-scale features, and a pixel-space loss countering latent compression. Coupled with learnable viewpoint embeddings, \ttt\ robustly supports both single- and multi-view inputs.
}
\label{fig:pipeline} 
\end{figure*}

\subsection{Data Construction Agent}
\label{sec:agent}

\paragraph{MLLM-Driven Procedural Geometry Synthesis.} 
We prompt a Multimodal Large Language Model (MLLM) to produce Blender scripts that generate target meshes (~\cref{fig:pipeline}). The candidate mesh is then rendered and visually critiqued by the MLLM to guide iterative code refinement, with an iteration cap enforced to control runtime. Once approved, the constructed geometry undergoes automated UV unwrapping. Owing to the simple topologies of our target geometries such as cans, cups and boxes, this pipeline is able to autonomously produce meshes with well-structured UV layouts, which is essential for subsequent texture generation.

\paragraph{UV-Guided Texture Synthesis and PBR Binding.} 
The UV layout mask acts as a spatial condition for an Image Generation Model (IGM) to synthesize textures in the UV domain, tasked with producing detailed text and patterns (~\cref{fig:pipeline}). To prevent perceptual artifacts, the MLLM evaluates the texture map by checking pattern sharpness and boundary alignment against the UV layout. The verified texture is bound to the geometric mesh alongside MLLM-prescribed PBR parameters (roughness and metallic), which yields high-frequency textured (HFT) 3D assets.

\paragraph{Notes.} We employ Qwen-3.8 as the MLLM and Qwen-Image-3.0-Pro~\cite{QWEN_AI} as the IGM. Although our agent natively supports end-to-end construction, in practice we prioritize efficiency: we first produce a fixed set of category-specific meshes, and subsequently synthesize a large volume of diverse textures over their UV layouts. This choice is driven by the fact that our task demands far greater texture diversity than geometric diversity. Constrained by the capabilities of current IGM models, the synthesized textures may not align flawlessly with the given UV layouts, but this discrepancy does not compromise the training of \ttt\ in any manner.

\subsection{Detail Preservation Scheme}
\label{sec:scheme}

\begin{figure}
\centering
\includegraphics[width=\linewidth]{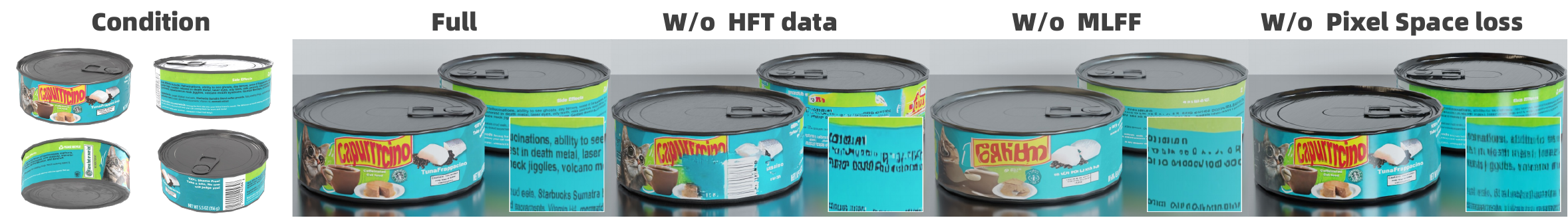}
\caption{\textbf{Effect of the HFT data, MLFF and pixel space loss.} The comparisons above validate the efficacy of each proposed component. Specifically, the HFT dataset is indispensable for reconstructing text and patterns; without it, the model fails to synthesize legible text. Furthermore, the MLFF module is crucial for faithfully reproducing textures from input images, while the pixel-space loss further sharpens subtle details, notably minute characters.}
\label{fig:ablation} 
\end{figure}

Using HFT data together with public datasets, we train a DiT model \( \mathcal{F}_{\theta} \) to recover PBR materials from a given image \( C \) and geometry \( \mathcal{G} \) with v-parameterization~\cite{salimans2022progressive}:
\begin{align}
\mathcal{L}_{\theta}
= \mathbb{E}
\left\|
\mathcal{F}_{\theta}\Bigl(\mathbf x_t, t, \mathbf z_c, \mathbf z_{\mathcal G}\Bigr)
- \mathbf v
\right\|_2^2,
\label{equ:training_loss}
\end{align}
where $\mathbf{x}_t$ is the latent feature at step $t$, and $\mathbf{z}_{\mathcal{G}}$ is the geometry latent from a pretrained encoder~\cite{xiang2025trellis2}. As deep semantic features $\mathbf{z}_c = \mathcal{E}(C)$ via DINOv2~\cite{oquab2024dinov2learningrobustvisual} fail to retain details (\cref{fig:ablation}), we design an MLFF module to overcome this.

\paragraph{Multi-level Feature Fusion.}

We extract DINOv2~\cite{oquab2024dinov2learningrobustvisual} features $\{\mathcal{E}^{(l)}(C)\}_{l=1}^{L}$ across $L$ layers spanning both shallow and deep levels from the image $C$. These features are aggregated into a single representation via $\mathcal{M}_{\phi}$ and jointly optimized with the denoiser $\mathcal{F}_{\theta}$:
\begin{equation}
\mathcal{L}_{\theta, \phi}
= \mathbb{E}
\left\|
\mathcal{F}_{\theta}\Bigl(\mathbf x_t, t, \mathcal{M}_{\phi}\bigl(\{\mathcal{E}^{(l)}(C)\}_{l=1}^{L}\bigr), \mathbf z_{\mathcal G}\Bigr)
- \mathbf v
\right\|_2^2.
\label{equ:training_loss2}
\end{equation}

We select $L=4$ layers (the 4th, 11th, 17th, and 23rd), which empirically suffices to recover diverse texture details. Furthermore, to enhance the spatial awareness of the fused representations, we augment them with learnable patch-level positional embeddings.

\paragraph{Latent-to-pixel Space Loss Transition.} 
Although computationally efficient, latent-space training inevitably incurs fidelity loss from compression artifacts (see w/o pixel space loss in ~\cref{fig:ablation}). To counter this, we introduce a coarse-to-fine supervision strategy that transitions to pixel space in late training stages. Formally, the predicted $\hat{\mathbf{v}}$ is mapped back to $\mathbf{x}_0$ and decoded into a PBR mesh, which is rendered into pixel-space albedo, roughness, and metallic maps across different viewpoints to provide multi-view photometric supervision:
\begin{equation}
\mathcal{L}_{\theta, \phi}^{*}
= \frac{1}{K}\sum_{i=1}^{K}
\left(
\lambda_1 \mathcal{L}_{\text{L1}}^{(i)}
+ \lambda_2 \mathcal{L}_{\text{SSIM}}^{(i)}
+ \lambda_3 \mathcal{L}_{\text{LPIPS}}^{(i)}
\right),
\end{equation}
where $K$ indicates the number of viewpoints under supervision. $\mathcal{L}_{\text{SSIM}}$ denotes the structural similarity loss~\cite{zhou2004ssim}, and $\mathcal{L}_{\text{LPIPS}}$ denotes the learned perceptual loss~\cite{zhang2018perceptual}. 
To preserve reconstruction capability, we supervise the views rendered from the same viewpoints as the input images with weights set to $(\lambda_1, \lambda_2, \lambda_3) = (1.0, 0.2, 0.2)$, favoring high photometric fidelity. Meanwhile, we render 2 novel viewpoints and adjust the weights to $(0.2, 0.2, 1.0)$ to enhance generative quality, emphasizing global perceptual realism over rigid per-pixel alignment.

\subsection{Multi-View Reconstruction}
\label{sec:multi}

\begin{figure}
\centering
\includegraphics[width=\linewidth]{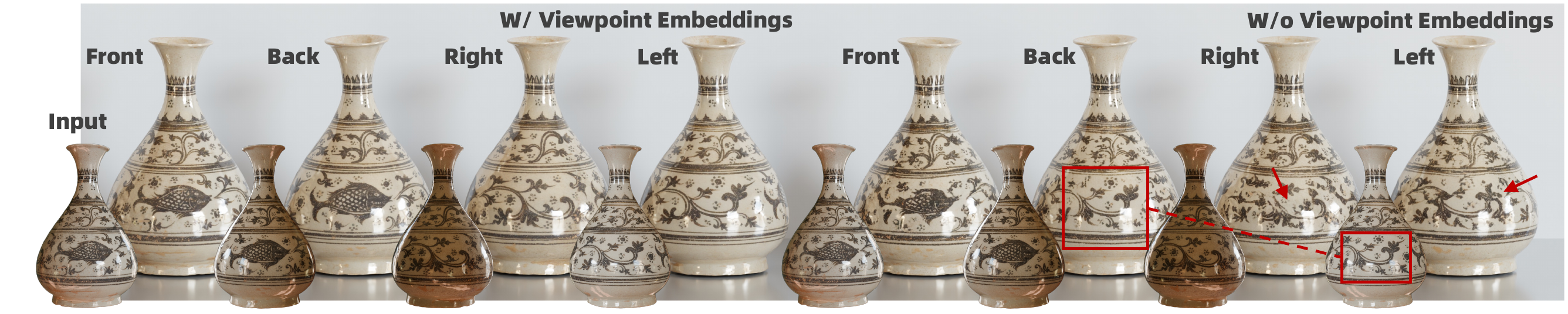}
\caption{\textbf{Effect of the viewpoint embeddings.} Red boxes highlight texture misplacement caused by omitting viewpoint embeddings (\textit{e.g.}, left-view textures are erroneously mapped to the back), while arrows indicate severe texture fragmentation in its absence.}
\label{fig:ablation_pe} 
\end{figure}

Incorporating inputs from additional viewpoints provides an alternative path to enhance texture fidelity. However, existing methods either lack native support for multi-view inputs or yield suboptimal results when generalized to this setting. To scale our pipeline to multi-view inputs, a straightforward approach is to concatenate the visual tokens of all input views before feeding them into $\mathcal{F}_{\theta}$. Nevertheless, the expanded token sequence exacerbates the difficulty of image-to-geometry feature alignment, leading to fragmented or incorrect texture generation (\cref{fig:ablation_pe}). 
We address this issue through a simple yet effective strategy: adding a learnable viewpoint embedding to the features of each view to disambiguate tokens originating from different viewpoints. These embeddings are jointly optimized with $\mathcal{F}_{\theta}$ and $\mathcal{M}_{\phi}$ to ensure seamless integration, and remain fixed during inference. Our model accommodates up to four input views, which proves sufficient for the vast majority of objects. These viewpoint embeddings correspond to canonical front, back, left, and right positions, covering a 90-degree horizontal range and a 180-degree vertical elevation range.

\paragraph{Optimizing via Geometric Ambiguity.}
Training on the HFT dataset is critical for properly optimizing the viewpoint embeddings. We intentionally curate this dataset with an abundance of symmetric primitives like cylinders and boxes, whose geometries are nearly indistinguishable across views. This structural ambiguity prevents the model from deducing viewpoints solely from shape cues, ensuring that the viewpoint embeddings are not bypassed. Meanwhile, rich multi-view appearance details compel the model to learn seamless cross-view transitions.

\paragraph{Low-SNR Timestep Sampling.} Multi-view texture reconstruction critically relies on the early denoising steps, which establish coarse texture localization before later steps refine details. This makes training near $t=1$ (pure noise) essential for learning the viewpoint embeddings. Motivated by this observation, we sample timesteps from a logit-normal distribution biased toward the low-SNR regime, with a mean of 2 and a standard deviation of 1.

\subsection{Implementation Details}

\paragraph{Dataset.} The overall training data is assembled from Objaverse-XL~\cite{deitke2023objaversexl}, TexVerse~\cite{zhang2025texverse}, and our HFT dataset, encompassing 500K, 300K, and 50K high-quality filtered PBR assets, respectively. Each asset is rendered across 16 viewpoints at a resolution of $1024\times1024$ under randomized fields of view and varied environmental lighting.

\paragraph{MLFF Architecture.} 
Multi-layer DINOv2 features are projected via dedicated two-layer MLPs and concatenated along the channel dimension, followed by an SE-inspired~\cite{hu2018squeeze} channel attention gate that adaptively reweights channels via global pooling and a two-layer gated excitation MLP. The calibrated features are then compressed along the channel dimension by a three-layer fusion MLP, added to the mean residual of all projected layers, and normalized through LayerNorm, achieving efficient cross-layer feature fusion with enhanced training stability. Across all aforementioned MLPs, intermediate hidden layers consistently employ GELU activations, while the excitation module terminates with a Sigmoid function to generate attention weights.

\paragraph{Optimization.} We fine-tune the material SC-VAE~\cite{xiang2025trellis2} on our dataset using 16 NVIDIA H20 GPUs (total batch size 64). Freezing the fine-tuned VAE, we train the DiT and the remaining modules under the same setup with a batch size of 16, using AdamW~\cite{loshchilov2017decoupled} with a learning rate of $1\times10^{-4}$. Training spans two 100K-step phases: first optimizing strictly with the latent-space loss, then continuing with the pixel-space loss, while alternating between single-view and multi-view inputs throughout.

\begin{table*}[t]
    \centering 
    \small
    \renewcommand{\arraystretch}{1.15}
    \setlength{\tabcolsep}{3.8pt}
    \caption{\textbf{Quantitative comparison under single-view input.} \textbf{Bold}: best, \underline{underlined}: second best.}
    \label{tab:quant_comp_grouped}
    \resizebox{\textwidth}{!}{
    \begin{tabular}{l
                    !{\vrule width 0.6pt}
                    ccc ccc
                    !{\vrule width 0.6pt}
                    ccc ccc}
    \toprule
        & \multicolumn{6}{c!{\vrule width 0.6pt}}{\textbf{Conditioning View Reconstruction}}
        & \multicolumn{6}{c}{\textbf{Novel View Generation}} \\
        \cmidrule(lr){2-7} \cmidrule(lr){8-13}
        & \multicolumn{3}{c}{Albedo}
        & \multicolumn{3}{c!{\vrule width 0.6pt}}{Rendering}
        & \multicolumn{3}{c}{Albedo}
        & \multicolumn{3}{c}{Rendering} \\
        \cmidrule(lr){2-4} \cmidrule(lr){5-7}
        \cmidrule(lr){8-10} \cmidrule(lr){11-13}
        Method
        & PSNR$\uparrow$ & SSIM$\uparrow$ & LPIPS$\downarrow$
        & PSNR$\uparrow$ & SSIM$\uparrow$ & LPIPS$\downarrow$
        & FID$\downarrow$ & CLIP-FID$\downarrow$ & CLIP-I$\uparrow$
        & FID$\downarrow$ & CLIP-FID$\downarrow$ & CLIP-I$\uparrow$ \\                                                                                                                                                                                         
    \midrule
    MaterialMVP~\cite{he2025materialmvp}
        & 18.90 & 0.830 & 0.160
        & 20.84 & 0.866 & 0.150
        & 124.10 & 17.12 & 0.8777
        & 102.88 & \underline{13.19} & 0.9037 \\

    UniTEX~\cite{liang2026unitex}                                                                                                                                                                                                                              
        & 21.84 & 0.873 & 0.101
        & 22.81 & 0.896 & 0.097
        & \underline{111.82} & 16.33 & 0.8896
        & 104.33 & 13.88 & 0.9074 \\

    Ink3D~\cite{han2026ink3d}
        & 19.39 & 0.837 & 0.155
        & 21.24 & 0.873 & 0.134
        & 143.89 & 21.08 & 0.8633
        & 112.94 & 14.88 & 0.9023 \\

    TRELLIS.2~\cite{xiang2025trellis2}                                                                                                                                                                                                                         
        & \underline{21.92} & \underline{0.879} & \underline{0.096}
        & \underline{23.46} & \underline{0.907} & \underline{0.090}
        & 113.86 & \underline{15.21} & \underline{0.9063}
        & \underline{87.69} & 13.21 & \underline{0.9150} \\

    \textbf{Ours}
        & \textbf{25.15} & \textbf{0.914} & \textbf{0.049}
        & \textbf{26.52} & \textbf{0.938} & \textbf{0.044}
        & \textbf{96.90} & \textbf{14.53} & \textbf{0.9096}
        & \textbf{77.66} & \textbf{12.11} & \textbf{0.9219} \\
    \bottomrule
    \end{tabular}}
\end{table*}

\begin{table}[t]
    \centering
    \scriptsize
    \caption{\textbf{Quantitative comparison under multi-view input.} Evaluated under four input views.}

    \begin{tabular}{
     l
     !{\vrule width 0.7pt}
     c c c
     !{\vrule width 0.7pt}
     c c c
    }
    \toprule 
    \multicolumn{1}{c!{\vrule width 0.7pt}}{} &
    \multicolumn{3}{c!{\vrule width 0.7pt}}{Albedo} &
    \multicolumn{3}{c}{Rendering} \\
    \cmidrule(lr){2-4} \cmidrule(lr){5-7}
     & PSNR$\uparrow$ & SSIM$\uparrow$ & LPIPS$\downarrow$
     & PSNR$\uparrow$ & SSIM$\uparrow$ & LPIPS$\downarrow$ \\
    \midrule
    MaterialMVP~\cite{he2025materialmvp}
    & \underline{20.18} & \underline{0.873} & \underline{0.117}
    & \underline{22.40} & \underline{0.900} & \underline{0.103} \\

    TRELLIS.2~\cite{xiang2025trellis2}
    & 19.91 & 0.852 & 0.162
    & 21.76 & 0.883 & 0.153 \\

    Ours
    & \textbf{25.48} & \textbf{0.919} & \textbf{0.051}
    & \textbf{27.30} & \textbf{0.941} & \textbf{0.045} \\
    \bottomrule
    \end{tabular}
    \label{tab:quant_multiview}                               
\end{table}

\section{Experiments}
\label{sec:experiments}

\paragraph{Baselines.} We compare \ttt\ with existing SOTA methods for evaluation. In the single-view setting, we compare against TRELLIS.2~\cite{xiang2025trellis2}, a 3D generation method, as well as view-space methods including UniTEX~\cite{liang2026unitex}, Ink3D~\cite{han2026ink3d}, and MaterialMVP~\cite{he2025materialmvp}. In the multi-view setting, we compare against TRELLIS.2, MaterialMVP, and ReconViaGen~\cite{chang2025reconviagen}, as the remaining baselines support only single-image conditioning. For TRELLIS.2, we implement the multi-view mode by switching the conditioning view during denoising, following TRELLIS~\cite{xiang2024structured}. To ensure a fair comparison of material generation, all methods use GT geometry except ReconViaGen, which cannot incorporate geometric input; therefore, we provide only qualitative comparisons for ReconViaGen. We also present qualitative comparisons with industry-leading commercial models.

\paragraph{Metrics.} As our method focuses on the generation of complex textures, we select 30 objects with rich textures from TexVerse~\cite{zhang2025texverse} for quantitative comparisons. We render views on both the conditioning and novel viewpoints of the generated objects to evaluate texture reconstruction and generation performance of all methods. For conditioning views, we report Peak Signal-to-Noise Ratio (PSNR), Structural Similarity Index Measure (SSIM)~\cite{zhou2004ssim}, and Learned Perceptual Image Patch Similarity (LPIPS)~\cite{zhang2018perceptual}. For novel views, we report Fréchet Inception Distance (FID)~\cite{heusel2017gans}, CLIP-based Fréchet Inception Distance (CLIP-FID)~\cite{radford2021learning}, and CLIP Image Similarity (CLIP-I).

\begin{figure*}[t]
\centering
\includegraphics[width=\linewidth]{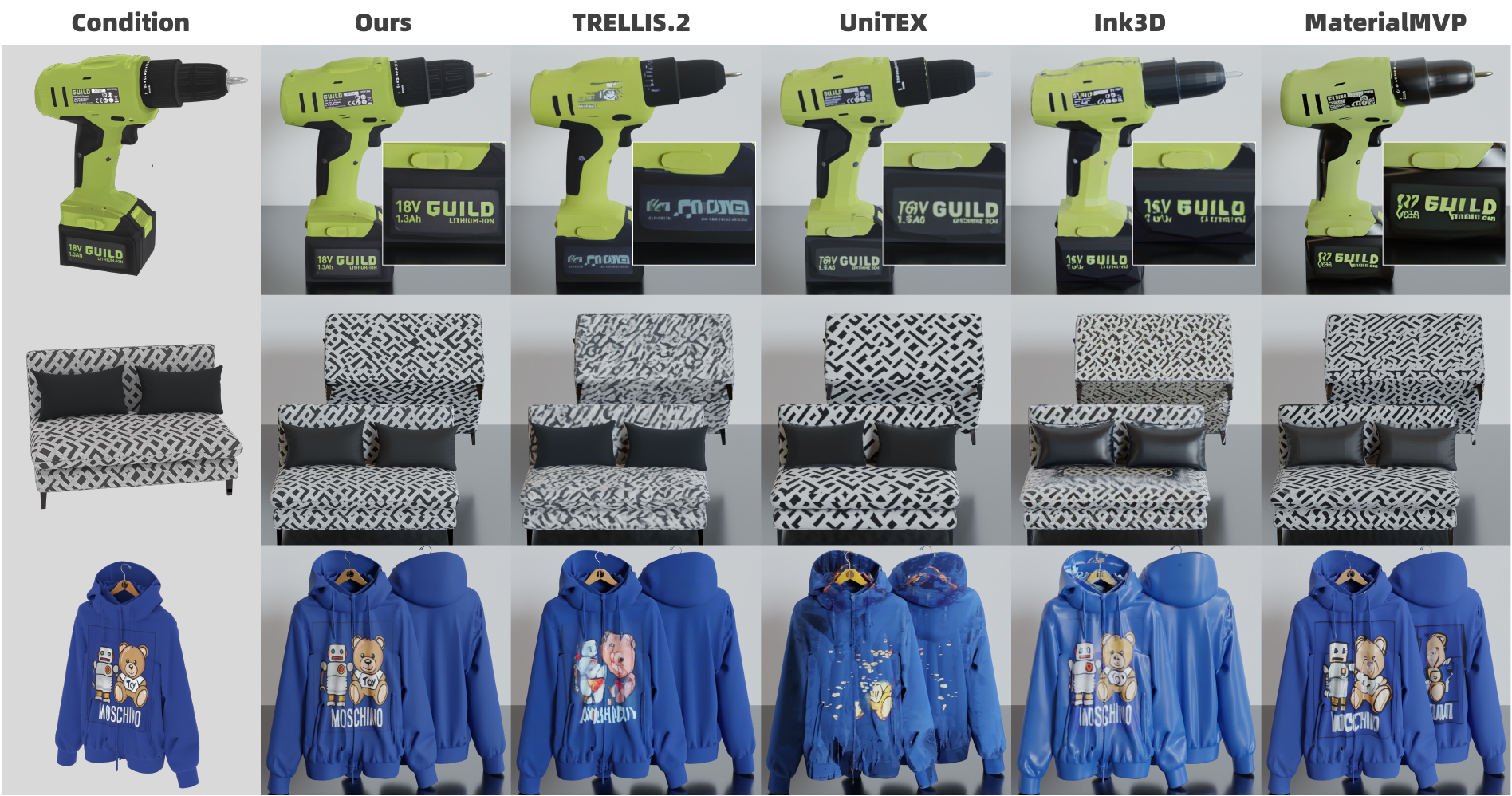}
\caption{\textbf{Single-view comparisons.} Our method achieves superior reconstruction fidelity, especially on text and patterns, while synthesizing highly coherent textures across unobserved regions.}
\label{fig:com_texverse} 
\end{figure*}

\begin{figure*}[t]
\centering
\includegraphics[width=\linewidth]{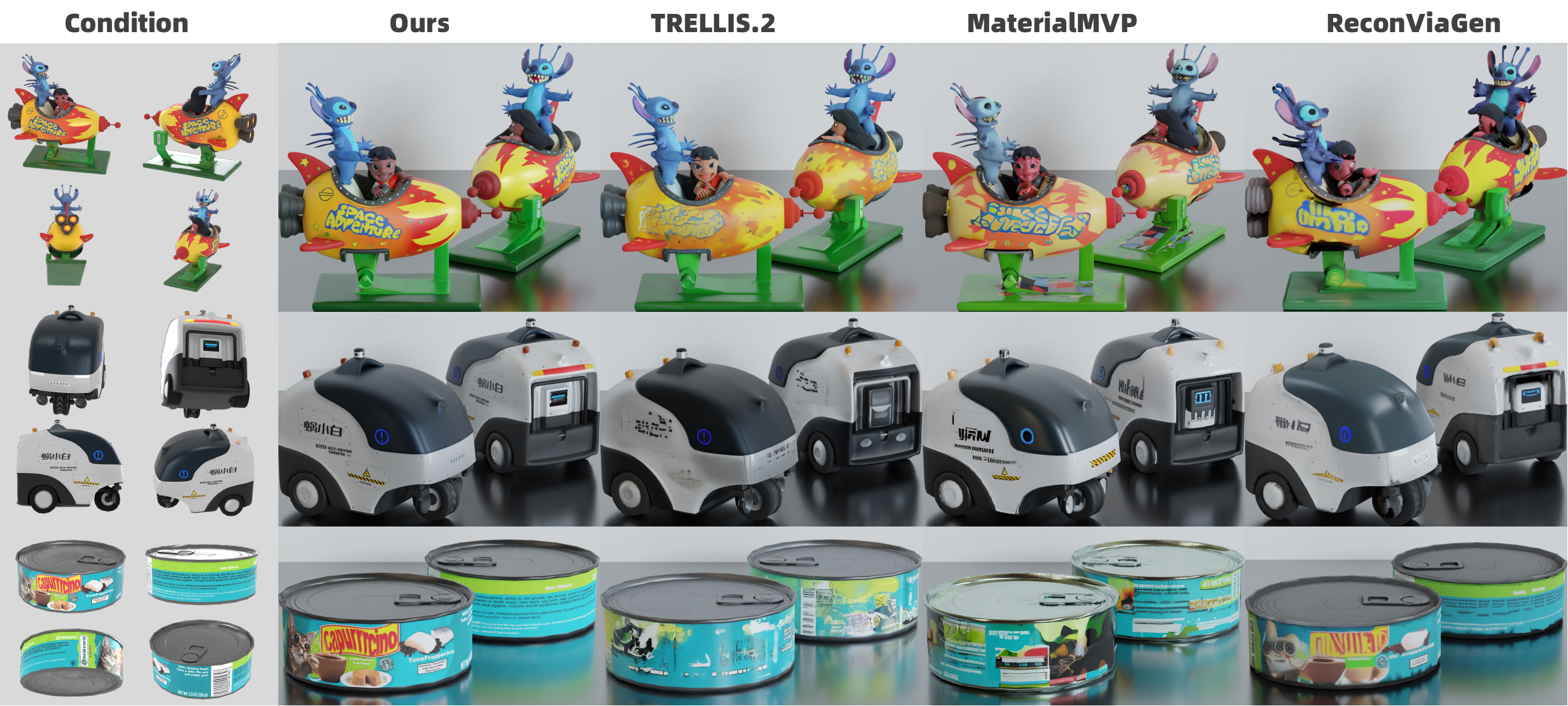}
\caption{\textbf{Multi-view comparisons.} Given four-view inputs, our method produces accurate, consistent and detailed textures, whereas competing methods yield erroneous and incoherent patterns}
\label{fig:com_texverse_multi} 
\end{figure*}

\begin{figure*}[t]
\centering
\includegraphics[width=\linewidth]{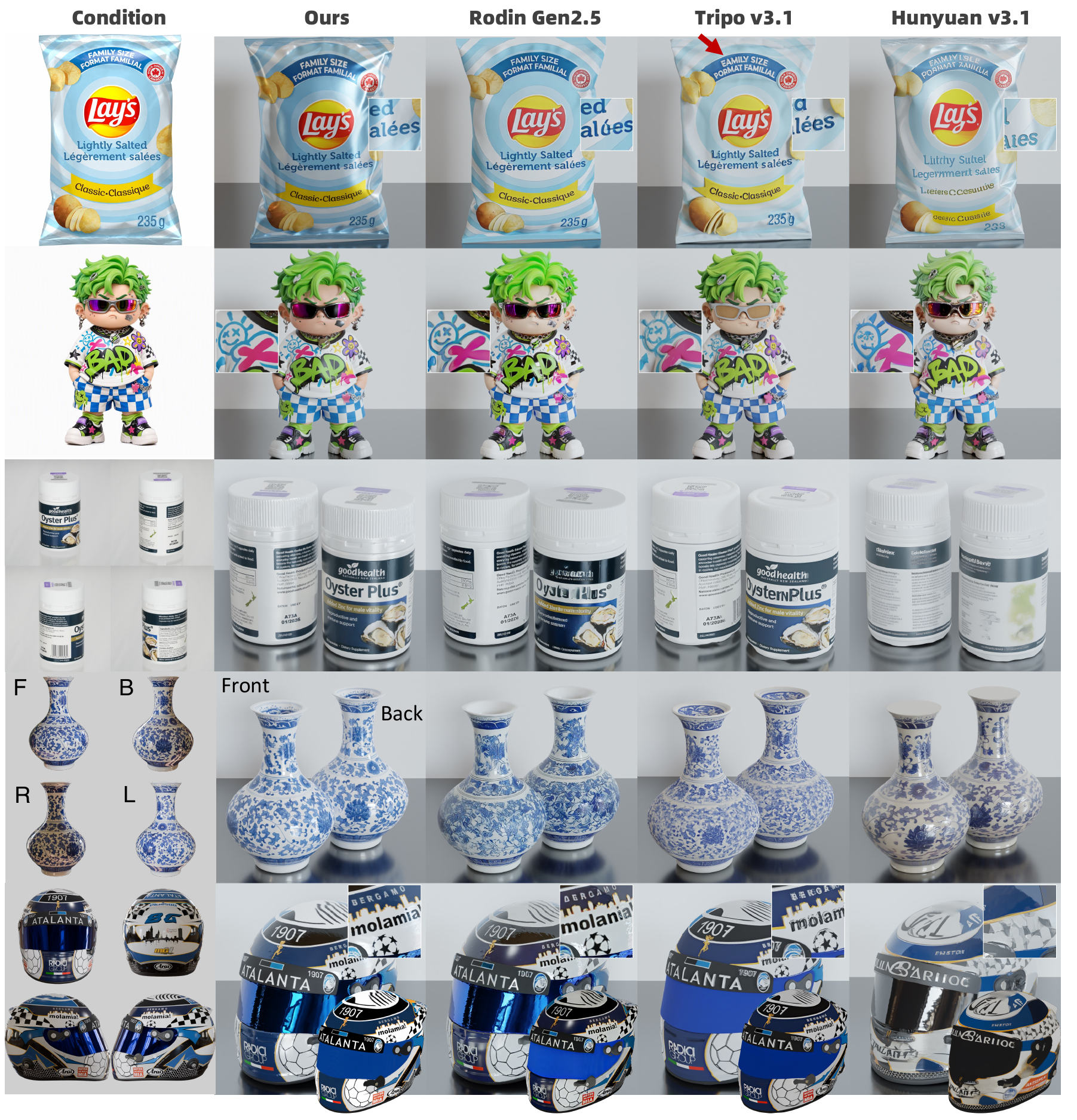}

\caption{\textbf{Comparisons with commercial models.} We use the geometry from Rodin Gen2.5~\cite{RODIN_AI} as our geometric input and others use their own geometries. Overall, \ttt\ achieves superior performance compared to existing commercial models in fine-grained reconstruction.}
\label{fig:com_texverse_commercial} 
\end{figure*}

\subsection{Comparisons}
\paragraph{Single-view Input.} 
Quantitative results in~\cref{tab:quant_comp_grouped} show that our method substantially surpasses all competing methods across metrics, confirming its advantage in material reconstruction and generation. Qualitatively (\cref{fig:com_texverse}), our approach faithfully preserves text and pattern details that other baselines fail to resolve. Among the alternatives, TRELLIS.2~\cite{xiang2025trellis2} yields coherent textures but lacks high-frequency details. UniTEX~\cite{liang2026unitex} leverages fine-tuned FLUX~\cite{labs2025flux1kontextflowmatching} to synthesize multi-view textures, yet suffers from instability and severe texture fragmentation. While Ink3D~\cite{han2026ink3d} and MaterialMVP~\cite{he2025materialmvp} achieve plausible visual quality, Ink3D suffers from noticeable geometry-appearance misalignment stemming from its video-based orbit synthesis, whereas MaterialMVP struggles to preserve fine-grained texture details, ultimately compromising their quantitative performance.

\paragraph{Multi-view Inputs.}
Our method continues to lead all baselines under multi-view conditioning (\cref{tab:quant_multiview}). TRELLIS.2 and MaterialMVP produce blurry and incorrect textures when conditioned on multiple views, and they are particularly unable to place textures at the correct locations on cylindrical cans (~\cref{fig:com_texverse_multi}), where geometric cues are less pronounced. ReconViaGen~\cite{chang2025reconviagen} reconstructs geometry and textures using VGGT~\cite{wang2025vggt} features. While it can generate roughly accurate textures, the results are often blurry and lack fine details. In contrast, benefiting from our well-learned viewpoint embeddings, our method accurately generates text and complex patterns on the correct geometric surfaces while producing sharp and consistent textures.

\paragraph{Comparison with Commercial Models.}
Fig.~\ref{fig:com_texverse_commercial} presents qualitative comparisons with commercial alternatives. In the single-view setting, our method faithfully preserves fine-grained text and patterns (e.g., on product packaging and clothes). In multi-view scenarios, whereas commercial models frequently suffer from texture misalignment and ghosting artifacts on symmetric objects (e.g., medicine bottles and vases), our approach accurately maps multi-view inputs onto corresponding surfaces with seamless transitions. Overall, our method outperforms competing commercial models in texture fidelity, material disentanglement, and generation stability.

\begin{table}[t]
    \centering
    \scriptsize
    \caption{\textbf{Ablation study.} All ablations are evaluated under the multi-view input setting.}

            \begin{tabular}{
             l
             !{\vrule width 0.7pt}
             c c c
             !{\vrule width 0.7pt}
             c c c
            }
            \toprule 
            \multicolumn{1}{c!{\vrule width 0.7pt}}{} &
            \multicolumn{3}{c!{\vrule width 0.7pt}}{Albedo} &
            \multicolumn{3}{c}{Rendering} \\
            \cmidrule(lr){2-4} \cmidrule(lr){5-7}
             & PSNR$\uparrow$ & SSIM$\uparrow$ & LPIPS$\downarrow$
             & PSNR$\uparrow$ & SSIM$\uparrow$ & LPIPS$\downarrow$ \\
            \midrule
            W/o HFT data
            & 22.90 & 0.896 & 0.087
            & 24.86 & 0.923 & 0.077 \\
    
            W/o MLFF 
            & 23.40 & 0.905 & 0.072
            & 25.34 & 0.928 & 0.065 \\
     
            W/o viewpoint embeddings
            & 24.47 & 0.911 & \underline{0.062}
            & \underline{26.47} & 0.935 & \underline{0.054} \\
     
            W/o pixel-space loss
            & \underline{24.61} & \underline{0.913} & 0.064
            & 26.22 & \underline{0.935} & 0.057 \\
     
            Ours
            & \textbf{25.48} & \textbf{0.919} & \textbf{0.051}
            & \textbf{27.30} & \textbf{0.941} & \textbf{0.045} \\
            \bottomrule
            \end{tabular}
            \label{tab:ablation}
\end{table}

\subsection{Ablation Study}

\paragraph{HFT Dataset.}The HFT dataset substantially enhances high-frequency details, particularly text patterns. As illustrated in~\cref{fig:ablation}, training with HFT data allows our model to faithfully recover clear text across curved surfaces. Quantitative results in~\cref{tab:ablation} further corroborate these gains in detail fidelity, validating both our data construction agent strategy and the effectiveness of the curated data.

\paragraph{Multi-level Feature Fusion.} ~\cref{fig:ablation} compares our fused features with semantic features alone. By aggregating multi-level representations, the fused features from MLFF better preserve fine-grained details from the input images. Without MLFF, the model struggles to reconstruct patterns and text, producing fragmented, blocky, and distorted textures. Coupled with the quantitative results in~\cref{tab:ablation}, this confirms that the MLFF module effectively enhances texture fidelity.

\paragraph{Pixel-space Loss.} Both~\cref{tab:ablation,fig:ablation} demonstrate that introducing the pixel space loss alleviates reconstruction errors from the material SC-VAE, thereby further refining fine details. Without this constraint, optimizing solely in the latent space produces visually plausible textures overall, but tiny text exhibits noticeable distortion (see the zoomed-in views in~\cref{fig:ablation}).

\paragraph{Viewpoint Embeddings.} The ablation results in~\cref{tab:ablation,fig:ablation_pe} validate the efficacy of our viewpoint embeddings and the associated training strategy. Without viewpoint embeddings, the model exhibits spatial ambiguity on nearly symmetric objects with repetitive, intricate patterns (e.g., the vase), often misplacing textures across surface regions. This positional mismatch consequently leads to distorted and fragmented textures (highlighted by the red arrows in~\cref{fig:ablation_pe}).

\section{Conclusion}
\label{sec:conclusion}

\ttt\ enables high-fidelity native 3D material reconstruction from single- or multi-view inputs via co-designed data, architecture, and training strategies. We leverage a designed data agent to create high-frequency assets, an MLFF module with pixel-space loss to maximize detail preservation, and viewpoint embeddings to ensure accurate multi-view reconstruction. Despite outperforming existing methods, \ttt\ remains constrained by voxel resolution when reconstructing minute text or patterns. Since higher resolutions demand prohibitive GPU memory, scaling resolution under a limited memory budget to improve details presents a compelling avenue for future research.

\clearpage

\bibliography{main}
\bibliographystyle{iclr2027_conference}

\end{document}